\pdfoutput=1

\documentclass[11pt]{article}

\usepackage[]{ACL2023}
\usepackage{graphicx}
\usepackage{times}
\usepackage{latexsym}

\usepackage[T1]{fontenc}

\usepackage[utf8]{inputenc}

\usepackage{microtype}

\usepackage{inconsolata}

\title{Check News in One Click:\\ NLP-Empowered Pro-Kremlin Propaganda Detection}

\author{Veronika Solopova \\
Technische Universität Berlin, Germany \\
 \texttt{veronika.solopova@tu-berlin.de} \\\And
 Viktoriia Herman \\
 Berlin, Germany \\
 \texttt{gvika739@gmail.com} \\
 \\\AND
Christoph Benzmüller \\
Universität Bamberg, Germany \\
 \texttt{christoph.benzmueller@uni-bamberg.de } \\
 \\\And
Tim Landgraf \\
 Freie Universität Berlin, Germany \\
 \texttt{tim.landgraf@fu-berlin.de } \\}

\begin{document}
\maketitle
\begin{abstract}
Many European citizens become targets of the Kremlin propaganda campaigns, aiming to minimise public support for Ukraine, foster a climate of mistrust and disunity, and shape elections \cite{links}. To address this challenge, we developed “Check News in 1 Click”, the first NLP-empowered pro-Kremlin propaganda detection application available in 7 languages, which provides the lay user with feedback on their news, and explains manipulative linguistic features and keywords. We conducted a user study, analysed user entries and models' behaviour paired with questionnaire answers, and investigated the advantages and disadvantages of the proposed interpretative solution.
\end{abstract}
\section{Introduction} 
Evidence that we are living through a global crisis of trust in news is substantial, which inspired many a debate concerning the measures needed to rebuild it \citep{Flew2020TrustAM, Gaziano1988-ys}. An increasing number of people are getting their news online, particularly the younger generation, while many have started avoiding the news, first those concerning the COVID-19 pandemic and now those about the Russian war in Ukraine, majorly due to low credibility and negativity.\citep{avoid}. At the same time, digital platforms are viewed more sceptically, than traditional news, especially political ones as they are believed to be agenda-driven and contain propaganda \citep{trustgap, Flanagin2000-qo,35countries}.\\ State-sponsored pro-Kremlin propaganda became a major issue, as reports claim that only a small per cent of Russian bots are being uncovered and detected \citep{only1percent}. \citet{geissler2023russian} showed that Twitter's (now X's) activity supporting Russia generated nearly 1 million likes, about 14.4 million followers and a substantial proportion of pro-Russian messages that went viral.\\ To address this issue, we created an accessible online user interface to check news in terms of pro-Kremlin propaganda, general manipulation and non-neutrality in 7 languages. It receives users' news and offers the model's verdict, its probability, as well as an explanation of manipulative keywords, linguistic strategies and indicators, shown to be associated with pro-Kremlin news. In addition to the models from our previous study \cite{Solopova2023-az}, we trained new ones for Italian and German languages, exploring the usefulness of the data-augmentation strategy through translation, as well as multi-language versus language-specific pre-trained transformer models for this task. Here, we present our system architecture and the user study we conducted, quantifying user satisfaction and desirable features and analysing user entries.
\subsection{Related Work}
Many tools have been developed to warn readers about fake news and “weaponize” them to understand the manipulative news better. 
An increasing amount of tools are based on automated text analysis and classification, mostly available only for English. \textbf{The Factual}\footnote{https://www.thefactual.com} is rating the credibility of the news each day using the site’s sourcing history, the author’s track record, and the diversity of sources in a news article as key features. \textbf{ClaimBuster}\footnote{https://idir.uta.edu/claimbuster/} is an online tool for instant fact-checking, allowing users to check the veracity of their texts, by searching for a fact-checked claim similar to user's input. The \textbf{Fake News Graph Analyzer} characterises spreaders in large diffusion graphs \citep{BODAGHI2021100182}. The \textbf{Grover} \citep{zellers2019grover} uses a fake news detection model, which takes on the language of specific publications to detect misinformation more accurately. \textbf{Bad News} \citep{doi:10.1098/rsos.211719,Basol-2020} is a gamified platform intended to build user understanding of the techniques and tactics involved in disseminating disinformation. They show that attitudinal resistance against online misinformation through psychological inoculation may reduce cultural susceptibility to misinformation.\\
Considering propaganda detection as a specific case of disinformation, only a few projects develop comprehensive interfaces accessible to the public. \textbf{Proppy} \citep{BarrnCedeo2019ProppyAS} was trained on known propaganda sources using a variety of stylistic features and is constantly clustering news sources. \textbf{PROTECT} \citep{Vorakitphan2022-yw} and \textbf{Prta} \citep{prta} allow users to explore the articles, texts and URLs by highlighting the spans in which propaganda techniques occur through a dedicated interface. \textbf{Hamilton 2.0}\footnote{https://securingdemocracy.gmfus.org/hamilton-dashboard} is a real-time dashboard, created by the project of the Alliance for Securing Democracy, which aggregates analysis of the narratives and topics promoted by Russian, Chinese, and Iranian government officials state-funded and state-linked media accounts and news.
\textbf{NewsGuard} \footnote{https://www.newsguardtech.com/special-reports/russian-disinformation-tracking-center/} uses a team of journalists and experienced editors to produce reliable ratings and scores for news and information websites. To the best of our knowledge, no research-based open-source tools using AI to check potential Russian propaganda in a user’s specific piece of news and in several languages are currently available.
\section{Methods} 
\subsection{Data}
In addition to English, Russian, Ukrainian, French and Romanian, from our previous study, we chose to add German and Italian models to our tool. According to the European Union project EUvsDisinfo\footnote{https://euvsdisinfo.eu}, “no other EU member has been subjected to such a powerful disinformation attack as Germany has been”. In its database of fake media pieces accumulated since late 2015, German media holds the 1st place, while Italy is in third.\\ We used fact-checked and attested pro-Kremlin propaganda articles from Propaganda Diary \citep{DB}. Around 5\% was also added from the press of political parties associated with pro-Kremlin sympathy. This amounted to 963 articles. As an example of trust-worthy media, we used VoxCheck's “white list” including sources such as ZDF, Der Welt, Frankfurter Allgemeine Zeitung, and Spiegel (676 altogether). As an augmentation set, we translated 537 neutral news with BBC and The Guardian translations from English to German using translators python API\footnote{https://pypi.org/project/translators/} and 565 RT.ru and Ria.news from Russian to German. Together native news set consists of \textbf{1639} texts, while the the augmented one is \textbf{2741}.\\
In Italian, we collected \textbf{2229} news from the Propaganda Diary, out of them 922 with attested Russian Propaganda and 1307 ones from the  “white list” (e.g. Internazionale, La Repubblica, Corriere). We augmented the ‘propaganda’ class by 304 samples with translations from Russian to Italian of Sputniknews, resulting in \textbf{2533} texts.
\begin{table}
\centering
\begin{tabular}{llll}
\hline
\textbf{Model} & \textbf{F1} & \textbf{MCC} & \textbf{AUC}\\
\hline
SVM-de-it & 0.82 & 0.64 & 0.82 \\
BERT-de-it & 0.01 & 0.51 & 0.51 \\
\hline
SVM-de-w/tr & 0.90 & 0.80 & 0.90 \\
SVM-de-w/o-tr & 0.92 & 0.83& 0.93 \\
BERT-de-w/tr & 0.94 & 0.88 & 0.99 \\
BERT-de-w/o-tr & 0.95 & 0.92 & 0.99 \\ 
\hline
SVM-it-w/tr & 0.78 & 0.57 & 0.78 \\ 
SVM-it-w/o-tr & 0.75 & 0.49 & 0.74 \\
BERT-it-w/tr & 0.96 & 0.80 & 0.96 \\
BERT-it-w/o-tr & 0.94 & 0.73 & 0.93 
\\\hline
SVM-multi & 0.88 & n/a & n/a \\
BERT-multi & 0.92 & n/a & n/a 
\end{tabular}
\caption{Evaluation of the models used in the study. MCC and AUC results are not given for SVM-multi and BERT-multi as in the previous study Cohen's kappa was used instead. The numbers are rounded to 2 digits after the comma. de- stands for German model, it- for the Italian, w/tr - with augmentation through translation, w/otr- without.}
\label{tab:metrics}
\end{table}
\subsection{Models}
The models from our initial study included one multilingual SVM model trained on morpho-syntactic features and keywords from the glossary of manipulative terms of Russian propaganda curated by the National Security and Defence Council of Ukraine and a fine-tuned multilingual BERT model \citep{DBLP:journals/corr/abs-1810-04805}. We followed a similar scheme for the German and Italian models. We first trained both Support Vector Machine (SVM) and BERT multilingual models for both languages together and augmented the data with translated articles. This approach only drew a 0.8 weighted F1-score for SVM and a drastically low 0.51 for the BERT model.
Training models separately increased performance in each language, except for the Italian SVM model (0.77 on average, and the highest score was 0.78.). The result for the German SVM increased to 0.87 in 5-fold cross-validation, and 0.9 on the best seed.
We used the bert-base-german-cased model and dbmdz pre-trained bert-base-italian-cased model, both implemented through HuggingFace\footnote{https://huggingface.co} framework.
The German model scored 0.94 F1, and 0.99 auroc, with 0.88 mcc, while the Italian one scored 0.90 F1, 0.93 auroc and 0.8 mcc on the best fold, with averages across the folds being 0.88 F1, 0,96 auroc, 0.77 mcc.\\
We decided to revise our augmentation policies and excluded non-native data. Interestingly, results dropped for both SVM and BERT models in the case of the Italian language (0.73 F1 on average) and drastic to 0.72 mcc, 0.94 auroc and 0.86 F1 averaged over 3 folds, although translations in the training set only accounted for 12\% of texts. In contrast, while translations were 40\% of the augmented set, the German model's performance slightly increased without them, with SVM achieving 0.91 F1 best and 0.89 on average and the BERT model gaining up to 0.036 in mcc and 0.1 in F1 (see Table \ref{tab:metrics} for training results). 
\begin{figure}[tp]
\centering 
\includegraphics[width=7cm]{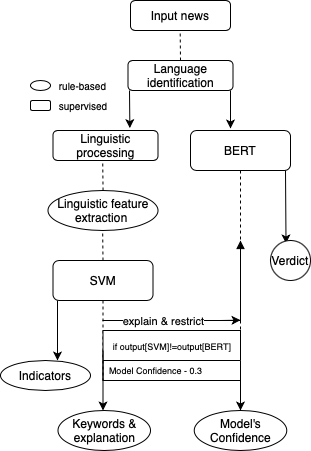}
\caption{The figure illustrates the system's mock-up. The elliptical elements are rule-based reasoners while squared ones are trained models.} \label{fig2}
\end{figure}
\begin{figure}[tp]
\centering 
\includegraphics[width=7.5cm]{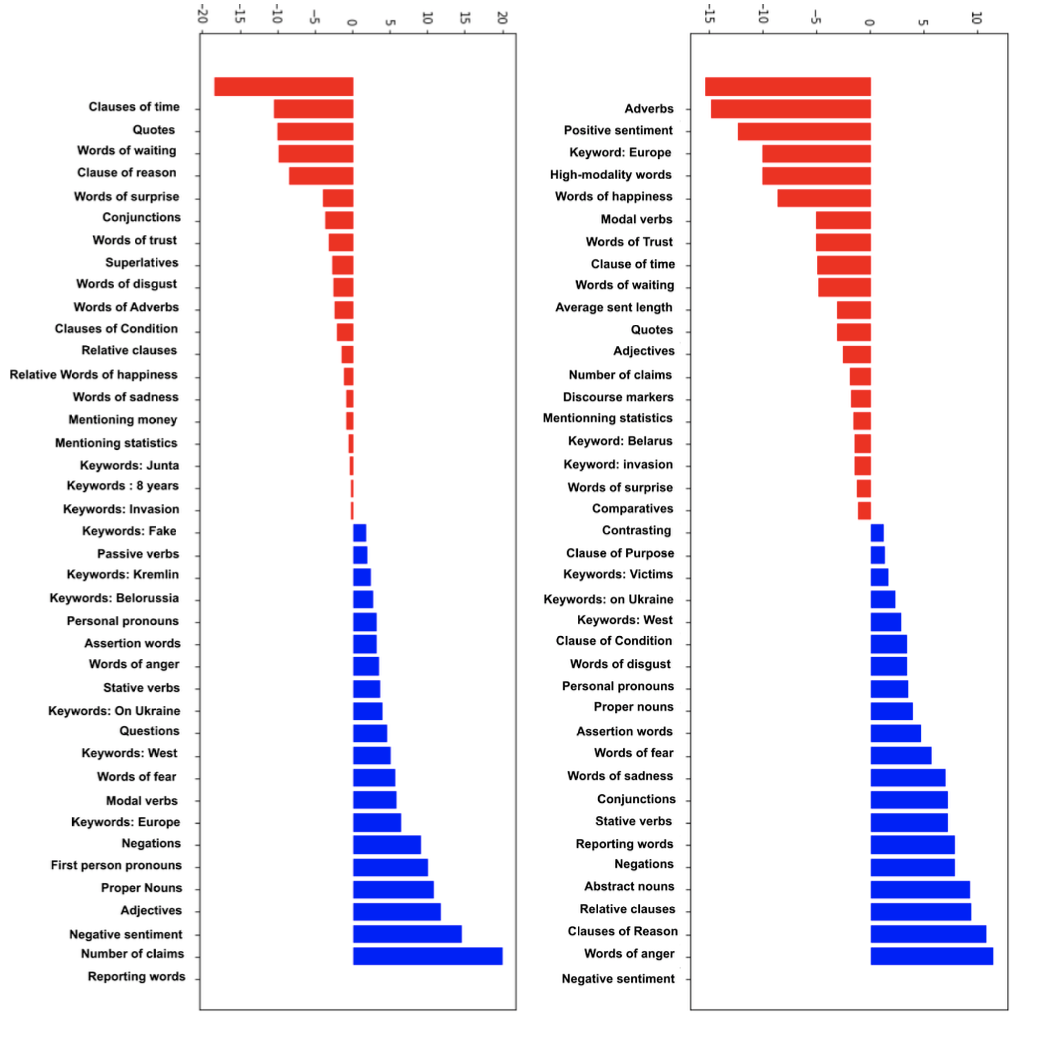}
\caption{The figure illustrates the distribution of the learnt features according to the stance. The upper red side shows the features with the highest negative coefficients for "Pro-Kremlin propaganda" prediction (hence, more likely in Western, Pro-Ukrainian media), while the lower blue side shows the coefficients indicative of "Pro-Kremlin propaganda".} \label{fig3}
\end{figure}
\subsection{System description}
\label{sysdes}
The interface is a web app, written with Python Flask framework for the back-end, and HTML, CSS and JavaScript for the front-end. The proposed news is fetched from the input window. The code for the front- and back-end is available under MIT License in our GitHub\footnote{https://github.com/verosol/propaganda\_website}.\\
First of all, the language is identified using langid.py \citep{lui-baldwin-2012-langid}. If the detected language is one of the languages we support the appropriate BERT model (language-specific for Italian and German and multi-language one for the rest of the languages) predicts the probability of propaganda in the text. If the text is longer than 520 tokens, it is divided into several chunks. If at least one contains propaganda, the whole text is classified as such. If the language is not supported, the news is translated into English using Traslators API. The program saves both the verdict, ‘Propaganda’ or ‘No propaganda’, and the probability of the predicted class. In parallel, the linguistic feature extraction script, using Spacy\footnote{https://spacy.io} for lemmatisation and part-of-speech tagging, analyses the whole body of the news and passes the feature and keyword vector to the specific SVM model (Italian, German or multilingual). If SVM predicts an opposite class from the BERT model, we deduct 45\% probability from the BERT's probability for the predicted class, and if the probability becomes lower than 30\%, we change the prediction to the opposite one. The mock-up can be seen in Figure \ref{fig2}.\\\\
For each RBF-kernel SVM model, we also trained a linear one and looked into the coefficients of features and keywords and their association with a particular stance (Figure \ref{fig3}). 
The top features are then used as linguistic indicators and are shown to the user as warnings of potential manipulative, non-neutral language associated with the stances. Important keywords are presented separately with explanations from the Glossary of the National Security and Defence Council of Ukraine on a click.
\begin{figure}[tp]
\centering 
\includegraphics[width=7cm]{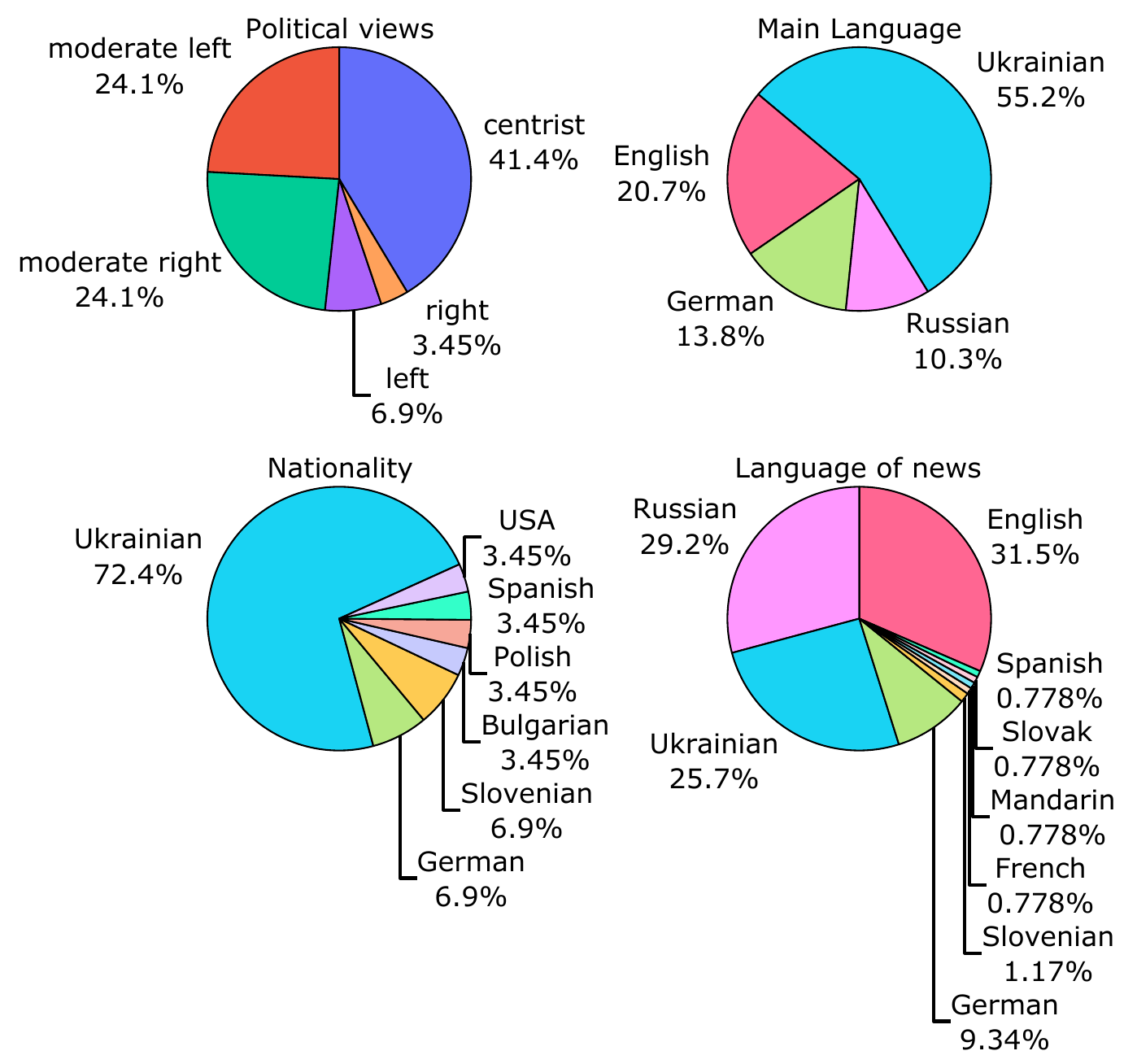}
\caption{The figure illustrates statistics on the users who took part in the survey and used the application.} \label{fig4}
\end{figure}
Comparing the important features and keywords, we discovered, that each language had its patterns of how Pro-Kremlin propaganda manifested itself, so we crafted indicators for each language separately. Some indicators, such as the abundance of negations, clause of purpose and reporting words, appeared to be universally indicative of pro-Russian propaganda in all of the languages we analysed. However, many features from our previous study, indicative of Pro-Kremlin propaganda were found more predictive of the Western stance in the two new languages. For example, frequent discourse markers, which are highly indicative of the pro-Kremlin side for other languages, are not associated with this prediction in German. The same stands for both German and Italian in terms of a high amount of quotes and clauses of time. In contrast, the clause of reason, highly predictive of a pro-Western stance for most languages, has the same tendency in Italian, but the opposite in German. 
\begin{figure*}[h]
\centering 
\includegraphics[width=\textwidth]{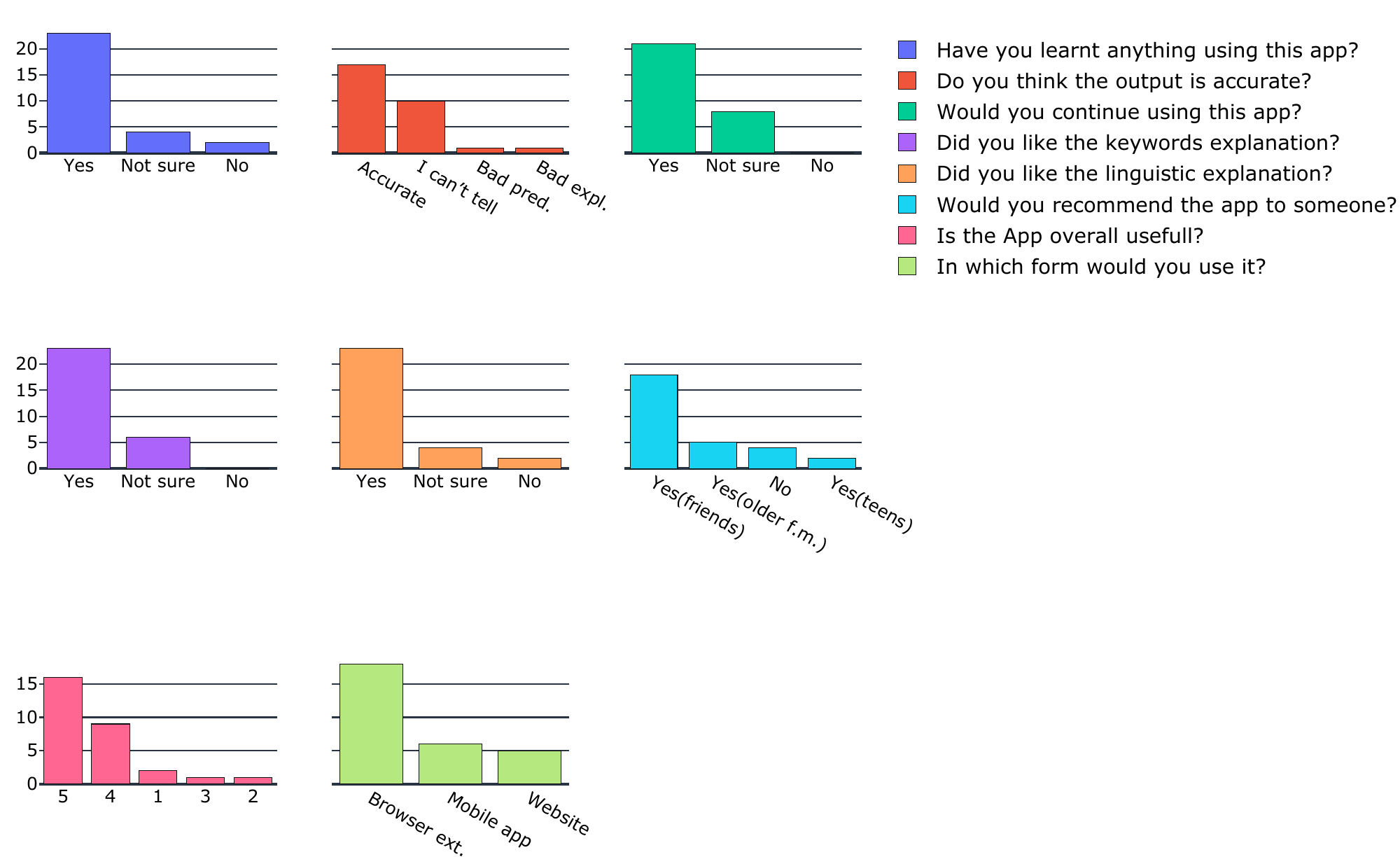}
\caption{The figure shows the results of the user study questionnaire.} \label{fig5}
\end{figure*}
\subsection{User study design}
Users were asked to check at least three different news in the app \footnote{https://checknewsin1.click} and fill out an integrated user questionnaire. \\ To understand the user profile we asked about the nationality, the language they searched in, their political stance, and how many pieces of news they verified. To quantify their experience, we asked their opinion about every element of the news analysis, its usefulness and accuracy, the preferable form (web application, desktop application, browser extension, chatbot), if they learnt something about propaganda and if they would continue using it, as well as the age group they would recommend this tool to (e.g. elder relatives, peers, teenagers, etc). From the back-end side, we collected the news the users entered, their own label (‘propaganda’ or not) and the analysis that the model provided.\\
The invitation to the user study was sent to various platforms on social media: several Italian, French and Ukrainian Facebook groups, subreddits r/EuropeanUnion, r/Samplesize, r/takemysurvey, r/YUROP,r/Ukraine, r/Ukraina; Dou.ua, a website for Ukrainian developers and IT workers, Instagram stories.
The user study contained the consent form. A system demonstration video\footnote{https://youtu.be/3dRXF5InGaE}  is available.
\section{Results}
191 users used the app with 257 unique requests, and only 29 out of them participated in the survey. 72\% of the users in the survey are of Ukrainian origin, central Europeans (Polish, Bulgarian, Slovenian, Slovak) account for another 15\%, 7\% German, with one American and Spanish user. Ukrainian was named as the main language only 55\% of the time though, while 20\% searched news in English, 13\% in German and 10\% in Russian.\\ The full pull of users showed further language variety: almost 1/3 of all news entered into the app were actually in English, 1/3 in Russian and a slightly smaller percentage in Ukrainian. Apart from 10 entries detected in German, other languages included French, Spanish, Slovenian and Mandarin. As for the political views of the respondents, 41\% self-identified as centrists, 24\% as moderate right or left, while only 7\% and 3.5\% were left or right respectively (see Figure \ref{fig4}).\\
\subsection{Survey results}
As illustrated in Figure \ref{fig5}, the majority of users (86\%) positively received the tool evaluating its usefulness as four or five on a scale of five, and only four respondents assessed the use as three and below. 79\% responded that they learned something new while using the tool. The same per cent liked the keyword explanations and linguistic indicators, whereas 72\% said that would continue using this app further. Only 58\% of users said that they think the output of the models was accurate while 34\% could not tell and 2 users either considered the verdict or the explanation to be wrong. 63\% would recommend the tool to their friends, 17\% to older relatives and only \~7\% to teenagers, while 13\% said they would not recommend it to anyone.\\
Talking about the potential formats for the tool, 62\% chose that browser extension would be the most preferable form, while mobile application is also slightly more preferred than the website option as it is (20\% against 17\%).
\subsection{User and model label comparison}
The multilingual BERT model showed an imbalanced prediction rate for different languages. The new German model had almost 50/50\% positive/negative prediction rate, similar to the labels provided by the users. At the same time in Russian and Ukrainian language the verdict ‘propaganda’ was issued by the model only 8\% of the time, while in English it was 28\%. In contrast, the users labelled almost identical amounts of news as ‘propaganda’ and ‘not propaganda’ in English and Ukrainian, while in Russian 73\% of submissions were claimed to contain it. Overall, only 21\% of verdicts and user labels coincide in German, 36\% in Russian, almost 50\% in English and 52\% in Ukrainian.\\
Diving deeper into the differences between the proposed and predicted labels, in German, there is an almost equal percentage of mismatch (41\% model: ‘No’, user: ‘Yes’ and 37\% model: ‘Yes’, user: ‘No’). In other languages, the model is majorly predicting ‘no propaganda’. In the case of Russian, e.g. the model did not predict ‘propaganda’ any single time when the user would say otherwise, with a similar result in Ukrainian (1.5\%).
\begin{figure}[tp]
\centering 
\includegraphics[width=7cm]{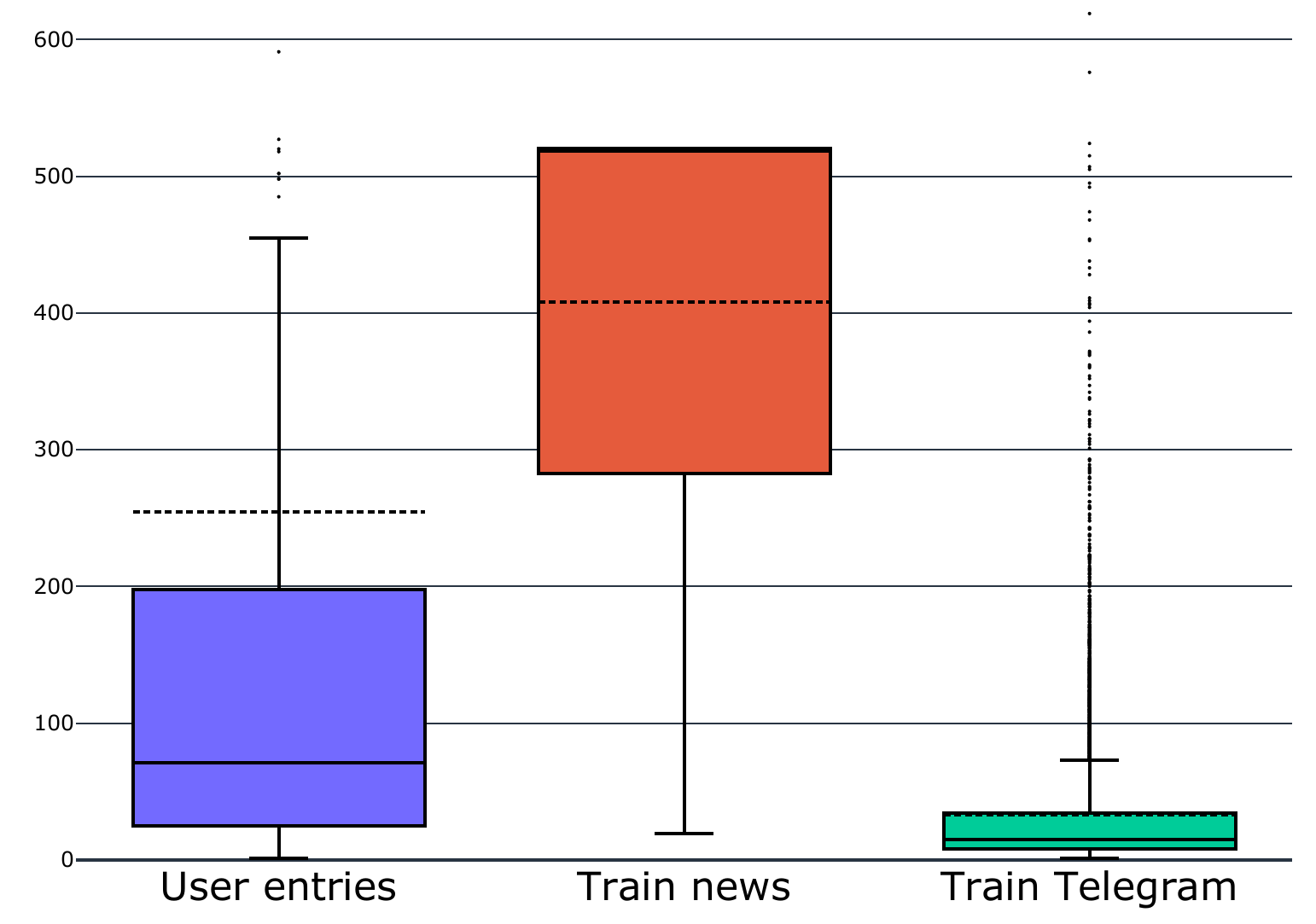}
\caption{The figure illustrates differences in the text length between the training sub-corpora and the user inputs.} \label{fig8}
\end{figure}
\section{Discussion and Conclusion}
Two major factors could explain the discrepancies between the user labels and the BERT predictions: either the user was wrong or the model, and here both tendencies seem to be present. As illustrated by Figure \ref{fig8}, we compared the distribution of the text lengths of the conventional news (which are rather large $\sim$407 tokens), Telegram news (which are rather short $\sim$32 tokens) in the training set and the news offered by the user ($\sim$205). We could see that the latter distribution with all quartiles falls perfectly in between the 2 training set constituents. Generally, the news can be even larger than the ones in our training set. For instance, the average article length of The New York Times is 622 words and 516 for The Washington Post \citep{newspapercoverage}.\\
 A brief qualitative analysis shows that while many inputs are indeed news, they are also majorly Reddit comments, tweets, and user-generated words and sentences. We implemented the opportunity for the user to provide us with the link and not only copy-paste a text, which then we scrape using newspaper library\footnote{https://newspaper.readthedocs.io}. Some inserted a link to Elon Musk’s tweets, and while X cannot be scraped.
On very long entries, the model did not once predict ‘propaganda’ and coincided in this prediction with the user. It had at least 15\% better matching with user labels on very short samples, similar to Telegram posts in length, proving that length can indeed be a reason for some miss-classifications, when the user was correct. However, the length is only the surface description of the underlying genre missing from the training material: the users are not as interested in conventional news checking, as in flagging and quick discovery of bots and malicious actors in social media comments and tweets. A high number of Ukrainian participants and a high number of certain responses concerning the tool's accuracy also showed that users predominantly were sure of their ability to recognize propaganda, but were interested in ways of quickly eliminating it from the informational eco-sphere.
The indicators and keywords provided an important addition to the main model’s verdict. Not only did the constraints we introduced on the main model help mitigate strong language-related biases, but they also appeared to be more reliable, as they do not mismatch as often with user annotations. Only in 16\% of cases where there were more pro-Russian propaganda features found and 8\%, where no pro-Western features were reported at all, would the user consider it a ‘no propaganda’ sample. With the user label being ‘Russian propaganda’, there was only 12\% with more pro-western than pro-Kremlin indicators identified, and 7\% where no pro-Kremlin associated features were offered to the user. The strong performance of the indicators may have had a positive influence on the overall user evaluation of feedback’s accuracy.\\
 Users also often underestimate their knowledge of propaganda or are not very attentive when providing the label. While we received a lot of negative labels, the linguistic features indicate that most of the news pieces are not neutral. 37\% of the news which was strongly not neutral were attributed to the ‘no propaganda’ label by the users. Only 6\% of truly neutral entries were rightfully annotated as such, and 4\% of them were called propaganda.\\
Overall, the results of the user survey, however limited in number, are positive. Both accuracy, recommendation, and interest in continuing to use the app are majorly high and both keywords and linguistic explanations were appreciated. In the free form, where we asked the users what they would like to change, it was even suggested to put more stress on the explanations and take away the overall verdict, showing the percentage of propaganda present. Apart from minor front-end suggestions, such as more visual support and instructions, some users were indicating that there was news with a pro-Western stance which were citing the President of Russia, which contained propaganda, and such cases may have to be dealt with separately. For the same reasons, the field of fact-checking is moving from the direct text-to-label classification towards more fine-grained and multi-featured info-sphere-based prediction \cite{Grover2022PublicWM}. The need to introduce many constraints for the main model through other models in our study is also a reflection of this trend. Including the layer user and human moderators in the research should become standard practice, as it helps better understand the needs of the community and tailor future solutions accordingly.
\section*{Ethics Statement}
The demographics of our study, although include different nationalities, are still predominantly from Ukraine, and young adults (who are the usual users of the platforms we used to market the study), thus excluding younger and more senior groups. We were also not able to attract Romanian and Italian users, despite targeted marketing in their groups.
It is also important to state that open-source propaganda research also provides malicious actors with the means to counteract automated tools and adapt the style so that it is even more difficult to detect in the future. We still claim that it is even more crucial to educate the wider public about the instruments to verify the news they consume.
\bibliography{anthology,custom}
\bibliographystyle{acl_natbib}

\end{document}